\documentclass[letterpaper, 10 pt, conference]{ieeeconf}  % Comment this line out
\IEEEoverridecommandlockouts                              % This command is only
\usepackage{url}
\usepackage{multirow}

\title{\LARGE \bf
Feature Analysis for Assessing the Quality of Wikipedia Articles through Supervised Classification
}

\author{Elias Bassani$^{1}$ and Marco Viviani$^{2}$% <-this % stops a space
\thanks{$^{1}$E. Bassani is with the Department of Informatics, Systems, and Communication (DISCo), University of Milano-Bicocca, and Consorzio C2T, Milan, Italy {\tt\small elias.bassani at consorzioc2t.it}}%
\thanks{$^{2}$M. Viviani is with the Department of Informatics, Systems, and Communication (DISCo), University of Milano-Bicocca, Milan, Italy {\tt\small marco.viviani at disco.unimib.it}}%
}

\begin{document}

\maketitle
\thispagestyle{empty}
\pagestyle{empty}

%%%%%%%%%%%%%%%%%%%%%%%%%%%%%%%%%%%%%%%%%%%%%%%%%%%%%%%%%%%%%%%%%%%%%%%%%%%%%%%%
\begin{abstract}

Nowadays, thanks to Web 2.0 technologies, people have the possibility to generate and spread contents on different social media in a very easy way. In this context, the evaluation of the quality of the information that is available online is becoming more and more a crucial issue. In fact, a constant flow of contents is generated every day by often unknown sources, which are not certified by traditional authoritative entities. This requires the development of appropriate methodologies that can evaluate in a systematic way these contents, based on `objective' aspects connected with them. This would help individuals, who nowadays tend to increasingly form their opinions based on what they read online and on social media, to come into contact with information that is actually useful and verified.

Wikipedia is nowadays one of the biggest online resources on which users rely as a source of information. The amount of collaboratively generated content that is sent to the online encyclopedia everyday can let to the possible creation of low-quality articles (and, consequently, misinformation) if not properly monitored and revised. For this reason, in this paper, the problem of automatically assessing the quality of Wikipedia articles is considered. In particular, the focus is on the analysis of hand-crafted features that can be employed by supervised machine learning techniques to perform the classification of Wikipedia articles on qualitative bases. With respect to prior literature, a wider set of characteristics connected to Wikipedia articles are taken into account and illustrated in detail. Evaluations are performed by considering a labeled dataset provided in a prior work, and different supervised machine learning algorithms, which produced encouraging results with respect to the considered features.

\end{abstract}

%%%%%%%%%%%%%%%%%%%%%%%%%%%%%%%%%%%%%%%%%%%%%%%%%%%%%%%%%%%%%%%%%%%%%%%%%%%%%%%%
\section{Introduction}

The development of the Web and digital technologies have allowed to considerably reduce the costs of production and the breakdown of information while increasing the ease of access.
Web 2.0 technologies, in particular, have given everyone the chance to generate and spread content online, in most cases without the intermediation of any traditional authoritative entity in charge of content control \cite{Eysenbach2008}. This augments the probability for people to incur into misinformation (e.g., opinion spam, fake news, \ldots) \cite{viviani2017credibility}, or low-quality information \cite{batini2016data}.
In the online scenario, traditional methods to estimate information quality -- such as the scrupulous analysis of contents by experts -- have become impractical, due to the huge amount of new content that is generated and shared every day on the Web. Therefore, it is necessary to design scalable and inexpensive systems to automatically estimate the quality of the information diffused, based on `objective' evidence. In this way, it could be possible to support users in relying on `certified' information, in an era where people increasingly form their opinions based on what they read online and on social media \cite{PEW2018}.

One of the main sources of knowledge freely accessible and editable by users is today the online encyclopedia Wikipedia.\footnote{https://www.wikipedia.org/} The peculiarity of the platform, i.e., the fact that it allows anyone to create and modify articles, constitutes both a strength and a weakness: on the one hand, this encourages the collaborative construction of knowledge, but, on the other hand, this can lead to the possible generation of low-quality or biased articles. To overcome this problem, groups of volunteers periodically monitor the content of Wikipedia articles, but their limited number confronted with the articles growth rate do not allow an overall and constant control. Furthermore, the subjectivity connected to human assessors results in a different quality evaluation for different articles belonging to distinct topic areas. All these open issues have to be tackled for maintaining the authoritativeness of the platform.
In this context, the proposed approach aims at automating the classification of Wikipedia articles on qualitative bases, by employing supervised learning. The approach focuses, in particular, on the analysis of multiple hand-crafted features that can be employed by well-known machine learning techniques, some of which previously applied in the literature. An in-depth analysis has been performed both on the syntax, the style and the editorial history of Wikipedia articles, and on the Wikipedia classification process, highlighting very relevant aspects not previously treated in the literature. For evaluation purposes, a labeled dataset generated and made publicly available in a previous work has been employed \cite{bassani18}. %The promising results obtained confirm the effectiveness of the proposed solution.

\section{Background: Wikipedia}

Wikipedia is a collaborative encyclopedia, launched on January 15, 2001 by Jimmy Wales and Larry Sanger. The idea of a collaborative online encyclopedia is antecedent; in 1993, Rick Gates proposed this model under the name of Interpedia \cite{reagle2010good}. Wikipedia is to all effects a `spiritual' successor of this proposal. The philosophical concept at the bases of Wikipedia was proposed by Richard Stallman in December 2000, in contrast to the digital encyclopedias then existing, and it is based on the idea that no centralized organization should have control over the editing process.\footnote{https://en.wikipedia.org/wiki/History\_of\_Wikipedia} On Wikipedia, users, besides being able to freely read articles, can also modify them and create new ones. This poses the responsibility for the quality of contents totally in the hands of those who contribute to them. From a purely theoretical point of view, the collaborative aspect of Wikipedia should avoid, on the one hand, the prevalence of an individual's point of view on those of others, and, on the other hand, bias and manipulation, since contents are subjected to control and revision by a large number of people from all over the world, resulting in a high quality of articles. Unfortunately, from a practical point of view, things are not so simple.
%
%From a technological point of view, Wikipedia is based on the \emph{wiki} technology, proposed in 1995 by Ward Cunningham \cite{leuf2001wiki}. A wiki is a Website that provides the possibility to modify its content and structure in a collaborative way directly from the Web browser.\footnote{https://www.britannica.com/topic/wiki} For collaboration, in this case, it is meant a \emph{modus operandi} that allows anyone, anywhere, to contribute to the contents of the platform, either by modifying them directly, or by participating in the discussions that concern them.
%
%\textcolor{blue}{HO AGGIUNTO DELLE INDICAZIONI TEMPORALI E CAMBIATO ALCUNE COSE SOTTO}
%

According to Alexa Internet, a company that analyzes Web traffic, Wikipedia is the fifth most visited Website in the world in 2018.\footnote{https://www.alexa.com/topsites} In the first six months of 2018 more than 250 million English Wikipedia pages were visited on a daily basis.\footnote{https://tinyurl.com/yb5jfzdq} As of August 2018, the English version of Wikipedia is constituted by 5.7 million articles and 650 new articles are created per day, while more than 3 million changes are made to existing articles monthly.\footnote{https://stats.wikimedia.org/EN/ReportCardTopWikis.htm\#lang\_en} It is therefore impossible for the 1,200 administrators of the English version to monitor the entire publishing activity carried out by editors.\footnote{https://en.wikipedia.org/wiki/Special:Statistics}
The huge flow of new information that everyday characterizes the publishing activity on Wikipedia, has the defect of introducing into the platform a huge number of just sketched or low-quality articles, whose utility for the users appears to be doubtful and could also cause, in the worst case, the proliferation of misinformation among the less attentive readers. To cope with this problem and to indicate to those who contribute to the platform the \emph{qualitative status} of an article, the Editorial Team of Wikipedia has defined some characteristics that an article should have in order to be considered of good quality,\footnote{https://en.wikipedia.org/wiki/Wikipedia:Content\_assessment\#Grades} and distinct \emph{quality classes} in which each article can be categorized based on its characteristics.

\subsection{The Wikipedia Quality Grading Scheme}\label{sec:QualityClasses}

Wikipedia is characterized, within its community, by groups of contributors, called \emph{WikiProjects}, which are focused on improving the articles belonging to particular topic areas (e.g., \emph{Mathematics}, \emph{History}, etc.). Today, the English version of Wikipedia has more than two thousand of these groups.\footnote{https://en.wikipedia.org/wiki/Wikipedia:WikiProject} Within each WikiProject, a so-called \emph{assessment team} deals with the evaluation of the quality of the articles. The qualitative evaluation of an article is, in line with the philosophy of the platform, an activity that every contributor can perform. According to Wikipedia,\footnote{https://en.wikipedia.org/wiki/Wikipedia:Content\_assessment} ``generally an active project will develop a consensus, though be aware that different projects may use their own variation of the criteria more tuned for the subject area''.
The evaluations provided by contributors are used primarily for internal uses of the project, %\footnote{\url{https://en.wikipedia.org/wiki/Wikipedia:WikiProject_Council/Assessment_FAQ\#Purpose}}
but they can also be accessed by readers to verify that the quality of an article is high before trusting its content. %Moreover, as mentioned in the previous section, some indications are provided directly on the main page of the article.

This evaluation relies on the WikiProject article quality grading scheme,\footnote{https://en.wikipedia.org/wiki/Template:Grading\_scheme} which divides the articles into seven distinct categories: $(i)$ \emph{Featured Articles}, denoted as FA-Class (FA) articles; $(ii)$ A-Class (A) articles; $(iii)$ \emph{Good Articles}, denoted as GA-Class (GA) articles; $(iv)$ B-Class (B) articles; $(v)$ C-Class (C) articles; $(vi)$ Start-Class (\emph{Start}) articles; $(ii)$ Stub-Class (\emph{Stub}) articles. The FA-Class includes the best articles on the platform, i.e., those considered complete and exhaustive from every point of view. In contrast, the Stub-Class includes all those articles that have a very basic description of the topic they deal with, or which are of very low quality. Intermediate classes are quality decreasing compared to the order in which they were previously listed.

Unfortunately, the number of new Wikipedia articles created every day and the number of changes made to existing ones makes it impossible the monitoring of their content by a small group of people, and the frequent verification and update of the quality classes they belong to. %This situation makes it necessary the design of models through which automate the classification of articles with respect to their quality.
To overcome this issue, over the years, several works tackling the problem of \emph{automatically} categorizing Wikipedia articles with respect to the above-mentioned quality classes have been proposed. The are illustrated in the following section.

\section{Background and Related Work}\label{sec:RelatedWorks}

The relevance of the information quality problem, the widespread use of Wikipedia and its collaborative nature have stimulated different studies aimed at automatically identifying the quality of its articles. The majority of these studies have addressed this issue as a classification task, and have proposed different approaches to categorize Wikipedia articles into quality classes. The first approaches employing machine learning algorithms to perform classification, inferred evidence of the quality of articles only by considering text features, i.e., features connected to the length of the text \cite{blumenstock2008size}, the language usage \cite{lipka2010identifying}, or some lexical aspects \cite{xu2011measuring}. Other works have proposed graph-based models to estimate a quantitative value representing the quality of an article \cite{hu2007measuring,sicilia2006evaluating, li2015automatically}. These models consider and combine different metrics related to both: $(i)$ the graph representing the editorial process of the articles, highlighting the relationships (edges) between articles and editors (nodes); $(ii)$ the graph representing links (edges) among articles (nodes), i.e., the Wikipedia articles graph. In general, the models proposed within this group evaluate both authors authority and articles quality. Another category of approaches employs (supervised) machine learning techniques acting on multiple kinds of features, encompassing text features and other features related to the writing style, the readability level, the analysis of the article structure, and other network-related metrics, to perform classification \cite{hasan2009automatic,dalip2014quality,rassbach2007exploring,stvilia2005assessing}.

Recently, a few approaches based on the use of Deep Learning have been proposed \cite{dang2016quality, dang2017end}. These approaches have proven to be effective in classifying Wikipedia articles over quality classes. Despite this, they do not involve \emph{hand-crafted} feature analysis, while the aim of this paper is to study and investigate the impact that specific groups of features have in assessing information quality on Wikipedia.
The next section, which represents the core of the paper, is devoted to the description and the analysis of the features that have been considered.

\section{Feature Analysis}\label{sec:featureAnalysis}

The choice of the features used to represent the elements on which to perform an automatic classification, is a fundamental operation in the majority of data-driven approaches \cite{fontanarava2017feature}. With respect to the issue of classifying Wikipedia articles with respect to different quality classes, an initial work has been devoted to the study of previously employed features to represent Wikipedia articles. According to \cite{hasan2009automatic}, three main groups of features can be identified: \emph{Text Features}, \emph{Review Features}, and \emph{Network Features}. The former group includes features extracted directly from the text, the second derives them from the analysis of the writing process, and the latter from the analysis of the graph of the articles. In state-of-the-art approaches, the number of features employed was quite limited: the most complete study is \cite{hasan2009automatic}, in which 69 features were considered. The present work extends the number of features that can be used for the classification of Wikipedia articles based on their quality, coming to consider 264 features. This consistent number derives in large part from an in-depth analysis carried out on the syntax used by the authors, in order to delineate, as precisely as possible, the use of language and the way in which sentences are constructed, and, therefore, the stylistic characteristics of the text. %of the articles.
Furthermore, also the editorial history has been analyzed in depth, highlighting various aspects not yet considered in the literature, such as the contributions deriving from the changes made to the articles by occasional users. The set of features extracted from the graph of the articles was not expanded with respect to previous works.

In the following, the considered features are illustrated. Because of their high number, a synthetic description will be provided only when necessary, to explain the rationale behind their choice with respect to the problem considered. With respect to state-of-the-art features, new features introduced in this paper are indicated by an asterisk.

%Vengono ora elencate le caratteristiche utilizzate, indicando, ove sia logicamente evidente, la correlazione con il concetto di qualità degli articoli, così come definito da Wikipedia mediante le linee guida per l'operazione di classificazione, ed evidenziando quali sono state aggiunte e quali sono invece state ereditate dai lavori precedenti. \\
%\linebreak
%\textbf{NB:} Tutte le caratteristiche che presentano un asterisco alla fine del nome sono state sono state ideate, od utilizzate in questo ambito per la prima volta, nel lavoro qui descritto.

\subsection{Text Features}\label{sec:TextFeatures}

\textit{Text Features} are the characteristics of the articles that can be extracted directly from their text. These features allow to highlight different aspects of the articles, such as the writing style, the structure, and the lexicon used. For this reason, they can be further divided into four sub-categories: $(i)$ \textbf{\textit{Length Features}}, connected to some length aspects of the articles; $(ii)$ \textbf{\textit{Structure Features}}, capturing the way in which articles are structured (i.e., paragraphs, sub-paragraphs, etc.); $(iii)$ \textbf{\textit{Style Features}}, highlighting the writing style and, therefore, the choices concerning the structuring of sentences and the use of the lexicon in drafting the articles; $(iv)$ \textbf{\textit{Readability Features}}, indicating the \emph{degree of readability} of the articles, i.e., the minimum scholastic level that is necessary to understand their contents.

In this paper, the study of new features focused on this area in particular, as the articles are mainly written texts and the textual characteristics turn out to be those that require less time and computational resources to be extracted. Moreover, these characteristics are, apart from rare cases, applicable in any context where there is a need to classify written texts on the basis of their quality, i.e., they are platform-independent.

\subsubsection{Lenght Features}

The length of an article can be an indicator of its quality. In fact, a good-quality text in a mature stage is reasonably neither too short (incomplete topic coverage), nor excessively long (verbose content). Further, in Wikipedia, \textit{Stub} articles (draft quality) are short in the majority of the cases, reinforcing the correlation between length and quality \cite{hasan2009automatic}. In this work, the following features have been considered: $(1), (2), (3)$ \textbf{Character} \cite{stvilia2005assessing}\textbf{/Word} \cite{rassbach2007exploring}\textbf{/Sentence} \cite{hasan2009automatic} \textbf{count}: the number of characters (including spaces)/words/sentences in the text; in addition, the new feature (4) \textbf{Syllable count*} has been introduced, counting the number of syllables in the text.

\subsubsection{Structure Features}

This group of features focuses on the way an article is (well/badly) organized. According to the Wikipedia quality standards,\footnote{https://en.wikipedia.org/wiki/Wikipedia:Version\_1.0\_Editorial\_Team/\\Release\_Version\_Criteria} a good article must be reasonably clear, organized adequately, visually adequate, and point to appropriate references and/or external links. The considered structural features are listed in the following, where the majority of state-of-the-art ones come from \cite{hasan2009automatic}, unless otherwise indicated: (1), (2) \textbf{Section/Subsection count}: the number of sections/subsections in the article. The intuition behind these features is that a good article is organized in sections (e.g., \emph{Introduction}, \emph{Summary}, \emph{List of references}, and \emph{External links}) and subsections; (3) \textbf{Paragraph count*}: the number of paragraphs constituting the article. The intuition behind this feature is that, in a high quality article, the text of sections and subsections should be further subdivided to facilitate the operation of reading and understanding the topics covered; (4), (5) \textbf{Mean section/paragraph size}; (6), (7) \textbf{Size of the longest/shortest section}, expressed in characters; (8) \textbf{Longest-Shortest section ratio*}. These features are useful to detect unusual section organization of articles with empty or very small sections, which could indicate incomplete content and drafts. Furthermore, (9) \textbf{Standard deviation of the section size}; (10) \textbf{Mean of subsections per section}; (11) \textbf{Abstract size}, expressed in characters. Mature articles are expected to have an introductory section summarizing its content; (12) \textbf{Abstract size-Article Length ratio*}: an article presenting an abstract whose length is very similar to its total length is probably incomplete. Features $(4)-(12)$ focuses on the correct balancing of an article. Other structure features are: $(13)-(15)$ \textbf{Citation count/count per section/count per text length}: the number of citations in the article/in sections/with respect to the total length of the article. A good-written article provides a sufficient and balanced number of citations; $(16)-(18)$ \textbf{External link count} \cite{stvilia2005assessing}\textbf{/links per section/links per text length}: the same rationale behind features $(13)-(15)$; $(19), (20)$ \textbf{Image count} \cite{stvilia2005assessing}/\textbf{Images per section}: the number of images in the text and the ratio between the number of images and sections. Pictures contribute to make content clearer and visually pleasant; $(21)$ \textbf{Images per text length*}: the ratio between the number of images and the length of the article, expressed in number of sentences.

\subsubsection{Style Features}

Aim of these features is to capture the writing style of contributors, i.e., ``some distinguishable characteristics related to the word usage, such as short sentences'' \cite{hasan2009automatic}. Many of the style features reported here below have been employed in \cite{hasan2009automatic}. When possible, for each feature the fist work having proposed it will be also indicated. The considered style features are: %New style features proposed in this paper are indicated with an asterisk\textcolor{blue}{COME SOPRA}:
$(1),(2),(3)$ \textbf{Mean} \cite{xu2011measuring}/\textbf{Largest} \cite{rassbach2007exploring}/\textbf{Shortest* sentence size}: the average number of words per sentence/the number of words of the longest/shortest sentence;
$(4),(5)$ \textbf{Large/Short sentence rate} \cite{rassbach2007exploring}: the percentage of sentences whose length is ten words greater/five words lesser than the article average sentence length;
$(6),(7)$ \textbf{Question count} \cite{rassbach2007exploring}\textbf{/ratio*}: the number of questions in the article/the ratio between question count and the total number of sentences in the article;
$(8),(9)$ \textbf{Exclamation count*/ratio*};
$(10)-(17)$ \textbf{Number of sentences that start with a pronoun / an article / a coordinating conjunction / subordinating preposition or conjunction} \cite{rassbach2007exploring}\textbf{ / a determiner* / an adjective* / a noun*/ an adverb*};
$(18)-(25)$ \textbf{Number of sentences that start with a pronoun- / an article- / a coordinating conjunction- / a subordinating preposition or conjunction- / a determiner- / an adjective- / a noun- / an adverb-Sentence count ratio*}: these features are built by considering the ratio between the value of features $(10)-(17)$ and the total number of sentences that make up the article;
$(26)-(44)$ \textbf{Number of modal auxiliary verbs* / passive voices} \cite{rassbach2007exploring}\textbf{ / `to be' verbs* / different words* / nouns* / different nouns* / verbs* / different verbs* / pronouns} \cite{hasan2009automatic} \textbf{ / different pronouns* / adjectives* / different adjectives* / adverbs* / different adverbs* / coordinating conjunctions* / different coordinating conjunctions* / subordinating prepositions and conjunctions* / different subordinating prepositions and conjunctions*} in the whole article; features $(45)-(62)$, which is a whole new group of features with respect to the literature, are the same as the group $(26)-(44)$ but computed per each sentence constituting the article;
%
%$(45)-(62)$ \textbf{Number of modal auxiliary verbs*/passive voices*/'to be' verbs*/ different words} \cite{xu2011measuring} \textbf{/ nouns* / different nouns* / verbs* / different verbs* / pronouns* / different pronouns* / adjectives* / different adjectives* / adverbs* / different adverbs* / coordinating conjunctions* / different coordinating conjunctions* / subordinating prepositions and conjunctions* / different subordinating prepositions and conjunctions* per sentence};
%
$(63)-(80)$ \textbf{Ratio between the number of modal auxiliary verbs* / passive voices* / `to be' verbs} \cite{rassbach2007exploring} \textbf{/ different words} \cite{xu2011measuring} \textbf{/ nouns} \cite{xu2011measuring} \textbf{/ different nouns* / verbs} \cite{xu2011measuring} \textbf{/ different verbs* / pronouns* / different pronouns* / adjectives* / different adjectives* / adverbs* / different adverbs* / coordinating conjunctions* / different coordinating conjunctions* / subordinating prepositions and conjunctions* / different subordinating prepositions and conjunctions* and the total number of words in the article};
$(80)-(82)$ \textbf{Ratio between the number of modal auxiliary verbs* / passive voice count* / `to be' verb* and the total number of verbs in the article};
$(83)-(89)$ \textbf{Ratio between the number of different nouns} \cite{xu2011measuring} \textbf{/ different verbs} \cite{xu2011measuring} \textbf{/ different pronouns* / different adjectives* / different adverbs* / different coordinating conjunctions* / different subordinating prepositions and conjunctions* and the total number of different words in the article};
$(90)-(91)$ \textbf{Average number of syllables/characters per words};
$(92)$ \textbf{Top-$m$ most discriminant character trigrams} \cite{lipka2010identifying}: they unveil the preferences of the authors for sentence transitions, as well as the utilization of stop-words, adverbs, and punctuation;
$(93)$ \textbf{Top-$n$ most discriminant POS trigrams}: they unveil the preferences of authors in constructing sentences \cite{lipka2010identifying}. To compute $m$ and $n$ for features $(92)$ and $(93)$, the ${\chi}^2$ statistical method provided by the Python library \texttt{scikit-learn} has been employed.\footnote{http://scikit-learn.org/stable/modules/generated/sklearn.\\feature\_selection.chi2.html}

\subsubsection{Readability Features}

These features are numerical indicators of the \textit{US grade level},\footnote{https://en.wikipedia.org/wiki/Educational\_stage\#United\_States} i.e., the comprehension level that a reader must possess to understand what is debated in a piece of text. They were first used, to tackle the considered problem, in \cite{rassbach2007exploring}. They comprise several metrics combining word counts, sentences, and syllables. The intuition behind these features is that ``good articles should be well written, understandable, and free of unnecessary complexity'' \cite{hasan2009automatic}. The set of features considered includes:
$(1)$ \textbf{Automated Readability Index} \cite{senter1967automated};
$(2)$ \textbf{Coleman-Liau Index} \cite{coleman1975computer};
$(3)$ \textbf{Flesch Reading Ease} \cite{flesch1948new};
$(4)$ \textbf{Flesch-Kincaid Grade Level} \cite{ressler1993perspectives};
$(5)$ \textbf{Gunning Fog Index} \cite{gunning1952technique};
$(6)$ \textbf{L\"{a}sbarhets Index} \cite{bjornsson1968lesbarkeit};
$(7)$ \textbf{SMOG-Grade} \cite{mc1969smog};
$(8)$ \textbf{Dale-Chall Readability Formula*} \cite{chall1995readability}. This latter metric, not previously employed for the quality assessment of Wikipedia articles, has been designed to numerically evaluate the difficulty of understanding that a reader encounters when s/he reads a text in English.

\subsection{Review Features}\label{sec:ReviewFeatures}

These features take their name from the fact that they are extracted from the review history of each article, i.e., how many times and in which way the article has been modified. They can measure the degree of maturity and stability of an article, since no extensive corrections could indicate good-quality articles having reached a maturity level, while a lack of stability could indicate different kinds of controversies (e.g., with respect to neutrality, correctness, etc.). In the following list of review features, \emph{registered users} are those having an explicit user profile and a username, \emph{anonymous users} are those identified only by their IP address, and \emph{occasional users} are those who edited the article less than four times (they may belong to one of the two categories mentioned above).

The considered features are: $(1)$ \textbf{Age} \cite{rassbach2007exploring}: the age (in days) of the article. Very recent articles are not normally considered of very high quality since they usually go through a refinement process; $(2)$ \textbf{Age per review} \cite{hasan2009automatic}: the ratio between age and number of reviews. It is used to verify the average period of time an article remains without revision; $(3)$ \textbf{Review per day} \cite{hasan2009automatic}: the percentage of reviews per day, to verify how frequently the article has been reviewed; $(4)$ \textbf{Reviews per user} \cite{dondio2006calculating}: the ratio between the number of reviews and the number of users. This feature is useful to infer how much reviewed is an article when contrasted against the number of reviewers; $(5)$ \textbf{Reviews per user standard deviation} \cite{dondio2006calculating}: this feature is useful to infer how balanced is the reviewing process among the reviewers; $(6)$ \textbf{Discussion count} \cite{dondio2006calculating}: the number of discussions posted by the users about the article. This is useful to infer conflict resolution and teamwork dynamics; $(7)$ \textbf{Review count} \cite{lih2004wikipedia}: the total number of reviews. $(8)$ \textbf{User count} \cite{lih2004wikipedia}: the total number of unique users that have contributed to the article. More contributors an article has, more objective its content is supposed to be; $(9)-(11)$ \textbf{Registered* / anonymous* / occasional* user count}; $(12)-(14)$ \textbf{Registered / anonymous / occasional user rate*}: Percentage of registered / anonymous / occasional contributors; $(15)$ \textbf{Registered/Anonymous user ratio*}: the ratio between registered and anonymous contributors; $(16)-(18)$ \textbf{Registered \cite{stvilia2005assessing} / anonymous \cite{stvilia2005assessing} / occasional* review count} \cite{stvilia2005assessing}: the umber of reviews made by registered / anonymous / occasional users; $(19-22)$ \textbf{Registered* / anonymous* / occasional \cite{dondio2006calculating} review rate*}: the percentage of reviews made by registered / anonymous / occasional users; $(22)$ \textbf{Registered-Anonymous review ratio*}: the ratio between reviews made by registered users and anonymous users. The previous four features aims to highlight a possible qualitative difference based on the ratios between the total number of reviews an article has and the reviews made by registered, anonymous and occasional users. $(23)$ \textbf{Revert count} \cite{stvilia2005assessing}: the number of times an article has been taken to a previous state (review annulment); $(24)$ \textbf{Reverts count-Review count ratio*}: the ratio between reverts count and review count; $(25)$ \textbf{Diversity} \cite{stvilia2005assessing}: the ratio between the total number of contributors and the number of reviews; $(26)$ \textbf{Modified lines rate} \cite{hasan2009automatic}: the number of lines modified when comparing the current version of an article with three-months older version. This is a good indicator of how stable an article is; $(27)$ \textbf{Last three-months review count*}: the number of reviews made in the last three months. This feature could indicate that the content of an article is controversial, the article is about evolving events or it is in the beginning of its editorial process; $(28)$ \textbf{Last three-months review rate} \cite{dondio2006calculating}: the percentage of reviews made in the last three months; $(29)$ \textbf{Most active users review count*}: the number of reviews made by the most active 5\% of users; $(30)$ \textbf{Most active users review rate} \cite{dondio2006calculating}: the percentage of reviews made by the most active 5\% of users; $(31)$ \textbf{ProbReview} \cite{hu2007measuring}: this measure tries to assess the quality of a Wikipedia article based on the quality of its reviewers. Recursively, the quality of the reviewers is based on the quality of the articles they reviewed.

\subsection{Network Features}\label{sec:NetworkFeatures}

Network features are extracted from the articles graph, which is built by considering citations among articles. These citations can provide evidences of the popularity of the articles. In addition, a high-quality article is expected to be used as a reference point for articles dealing with interconnected topics. Extracting this kind of feature is particularly onerous, due the magnitude of the graph. For this reason, state-of-the-art features have been considered:
%
%
%The absence of new features in this category derives from the complexity of the analysis of the graph from which they are inferred, since its magnitude does not allow their extraction in a short time\textcolor{blue}{VOLEVO METTERE I NUMERI MA HO VISTO CHE NELLA TESI PARLAVO DI UNA DISCREPANZA TRA IL NUMERO DI ARTICOLI DI WIKIPEDIA SECONDO LE STATISTICHE E QUELLI TROVATI NEL DATASET USATO - PRESO DA WIKIPEDIA STESSA - ne parlo nella sezione 4.2 delle tesi "Nota sui database forniti da Wikipedia"}.
%
$(1)$ \textbf{PageRank} \cite{brin2012reprint}: the
\textit{PageRank} of an article, previously employed in \cite{rassbach2007exploring}; $(2)$ \textbf{In-degree} \cite{hasan2009automatic}: the number of times an article is cited by other articles; $(3)$ \textbf{Out-degree} \cite{dondio2006calculating}: the number of citations of other articles; $(4)-(7)$ \textbf{Assortativity in-in / in-out / out-in / out-out}: ``the ratio between the degree of the node and the average degree of its neighbors. The degree of a node is defined as the number of edges that point to it (in-degree) or that are pointed by it (out-degree)'' \cite{hasan2009automatic}; $(8)$ \textbf{Local clustering coefficient} \cite{watts1998collective}: it aims at evaluating if an article belongs to a group of correlated articles; $(9)$ \textbf{Reciprocity}: the ratio between the number of articles that cite a specific article and the number of articles cited by that article; $(10)$ \textbf{Link count} \cite{dondio2006calculating}: the number of links to other articles. It differs from \textit{out-degree} since it counts also links to articles that have not been written yet (\textit{red links});\footnote{https://en.wikipedia.org/wiki/Wikipedia:Red\_link} $(11)$ \textbf{Translation count} \cite{dondio2006calculating}: the number of versions of an article in other languages.
Features related to assortativity and clustering coefficient were proposed in \cite{benevenuto2008identifying, castillo2007know, dorogovtsev2013evolution} for \textit{spam detection} in Web pages. They were previously used in \cite{hasan2009automatic}.

\section{Experimental Evaluation}

In this section, the effectiveness of the considered features is evaluated by testing different supervised machine learning classifiers, by employing the labeled datasets introduced in \cite{bassani18} and detailed in the following.

\subsection{The employed dataset}\label{dataset_description}

In the literature, prior supervised approaches have classified the articles of Wikipedia with respect to a \emph{subset} of the quality classes considered in this article (illustrated in Section \ref{sec:QualityClasses}). In fact, none of the previous studies have referred to the current scale, for different reasons: $(i)$ when the study was made the proposed quality scale was different \cite{hasan2009automatic}; $(ii)$ the authors decided to simplify the classification task on purpose. In \cite{rassbach2007exploring} the Stub-Class (drafts) has not been considered because it was believed to be too trivial to discern articles belonging to that class. In \cite{xu2011measuring} the authors consider only \textit{Featured Articles} and Start-Class articles. In \cite{blumenstock2008size, lipka2010identifying, xu2011measuring} the articles have been classified only as \textit{Featured Articles} and \textit{Random Articles} (or \textit{non-Featured Articles}); $(iii)$ the approaches did not perform a `real' classification into quality classes, but they provided a ranking of articles with respect to their quality. The ranking produced by \cite{hu2007measuring} is supposed to reflect the hierarchy of the quality classes: ``the perfect ranking should place all FA-Class articles before all A-Class articles, followed by all GA-Class articles and so on'', while in \cite{li2015automatically} the ranking is intended to identify \emph{relevant} VS \emph{non-relevant} articles, i.e., \emph{Featured} VS \emph{non-Featured Articles}.

Therefore, it has been necessary to proceed with the construction of a new dataset, as illustrated in \cite{bassani18}, by selecting labeled articles from the seven considered quality classes directly from Wikipedia. In doing so, it has been necessary to consider that
the guidelines for the classification provided by the Wikipedia Editorial Team are generic, and need to be specialized/refined depending on the topic area of interest. %In fact, according to Wikipedia:\footnote{\url{https://en.wikipedia.org/wiki/Wikipedia:Content_assessment}} ``generally an active project will develop a consensus, though be aware that different projects may use their own variation of the criteria more tuned for the subject area''.
For example, a high-quality article discussing \emph{Photography} is expected to have more images than one of similar quality dealing with \textit {Computer Science}, while a \textit{History} article is supposed to contain more dates with respect to \emph{Technology} ones, which probably will contain more technical details. The linguistic register is another aspect that it is influenced by thematic areas: a high-quality \emph{Economy} article will be characterized by a more complex lexicon compared to the one employed in a \textit {Literature for children} article. %It is therefore expected that quality classifications provided by distinct WikiProjects are based on similar consideration within each WikiProject, but possibly different considerations across different WikiProjects \textcolor{blue}{UN PO' TRICKY QUESTA FRASE}.
%Moreover, an article can be provided with multiple quality labels by distinct WikiProjects if its content touches different topic areas.
%
%in most of them, labeled datasets containing articles belonging to different topic areas have been employed, not considering that ``generally an active project will develop a consensus, though be aware that different projects may use their own variation of the criteria more tuned for the subject area''.\footnote{\url{https://en.wikipedia.org/wiki/Wikipedia:Content_assessment}}
%
%thus not dealing with the above-described issues \cite{blumenstock2008size, xu2011measuring, stvilia2005assessing, rassbach2007exploring, hasan2009automatic}. These approaches focus on the development of supervised techniques able to classify each Wikipedia article with respect to quality classes disregarding its topic area (and, therefore, the WikiProject it belongs to).
%Other approaches have considered specific topic areas to build suitable datasets to train their models \cite{hu2007measuring,li2015automatically,lipka2010identifying}, but no sufficient details were provided about these datasets.
%
For this reason, different labeled datasets must be generated for different topic areas, as done in \cite{hu2007measuring,li2015automatically,lipka2010identifying}. %In this way, the features remain those defined in Sections \ref{sec:TextFeatures}, \ref{sec:ReviewFeatures}, and \ref{sec:NetworkFeatures}; what changes are the articles that make up each dataset.
In this paper, in order to be able to experimentally evaluate the effectiveness of the features illustrated in Section \ref{sec:featureAnalysis} through different supervised classification strategies detailed in Section \ref{sec:Experiments}, the labeled dataset generated in \cite{bassani18} for the topic area \emph{Military History} %,\footnote{\url{https://en.wikipedia.org/wiki/Category:Military_history}}
has been employed. %which was, at the time of writing, the one with the highest number of articles per quality class, i.e., 1,090 FA-Class articles, 564 A-Class articles, 4,049 GA-Class articles, 13,675 B-Class articles, 22,604 C-Class articles, 77,908 Start-Class articles, 50,281 Stub-Class articles. %\footnote{\url{https://en.wikipedia.org/wiki/Wikipedia:Version_1.0_Editorial_Team/Military_history_articles_by_quality_statistics}}
The dataset consists of 400 articles randomly selected for each quality class, for a total of 2,800 articles. The dataset is therefore balanced with respect to classes. The dataset is publicly available, and further details are included in the documentation associated with the dataset.\footnote{http://www.ir.disco.unimib.it/WikipediaDataset/}
%
%During the experimentation process, an aspect worthy of consideration has emerged, which apparently had not been previously taken into consideration in the literature. As said before, Wikipedia is a highly dynamic platform whose contents are constantly modified. Therefore, the possibility that the version of the articles constituting Dataset $(a)$ - recently gathered from Wikipedia - is not the same version on which the original classification made by the Military History WikiPoject was performed, is far from remote. In fact, after an in-depth analysis, it has emerged that the
%Since articles present in Dataset $(a)$ were classified a significant amount of time before (and many versions before) the recently gathered version, %. For this reason, an additional experiment aimed at evaluating the impact of the noise introduced in the learning process by the modifications of the articles undergone after their classification by the WikiProject team has been performed. Therefore,
%a second dataset, namely Dataset $(b)$, has been built, composed of the same articles as Dataset $(a)$, but containing the original classified versions of the articles, which have been obtained by analyzing their revision history. This aspect has apparently not been taken into consideration in the literature. Both datasets are made publicly available.

\subsection{Experiments and results}\label{sec:Experiments}

Two experiments are illustrated in this section. Each experiment tests eight different classifiers based on distinct supervised machine learning techniques: \textit{Decision Tree} (DT), \textit{K-Nearest Neighbors} (KNN), \textit{Logistic Regression} (LR), \textit{Naive Bayes} (NB), \textit{Random Forest} (RF), \textit{Support Vector Classifier} (SVC), \textit{Neural Networks} (NN) and \textit{Gradient Boosting} (GB). In each experiment, the performance of classification has been evaluated in terms of \textit{Accuracy} and \textit{Mean-Squared Error} (MSE) \cite{kubat2015introduction}.

\subsubsection{Experiment 1}

In the first experiment, a comparative evaluation has been performed between the proposed approach and the state-of-the-art approach described in \cite{hasan2009automatic}, which employed supervised classifiers acting on the higher number of hand-crafted features among prior works in the literature. This experiment allows to evaluate the effectiveness of the features analyzed in this paper and those proposed by the baseline in classifying Wikipedia articles with respect to the seven quality classes (i.e., FA-Class, A-Class, GA-Class, B-Class, C-Class, Start-Class, and Stub-Class).

As reported in Table \ref{exp1acc} and Table \ref{exp1mse}, the set of features analyzed in this paper in conjunction with Gradient Boosting allow to obtain the best results in terms of both Accuracy ($61.8\%$) and MSE ($0.919$), with an improvement in terms of both measures with respect to \cite{hasan2009automatic}.%, where the best result is obtained by using the Random Forest classifier.

\begin{table}[htbp]
\begin{center}
\begin{tabular}{|c|c|c|}
\hline
\textbf{Classifier} & \textbf{Proposed Approach} & \textbf{Baseline \cite{hasan2009automatic}} \\
\hline
\textbf{DT} & 0.474 & 0.484 \\
\hline
\textbf{KNN} & 0.424 & 0.417 \\
\hline
\textbf{LR} & 0.497 & 0.499 \\
\hline
\textbf{NB} & 0.304 & 0.310 \\
\hline
\textbf{RF} & 0.592 & 0.601 \\
\hline
\textbf{SVC} & 0.506 & 0.539 \\
\hline
\textbf{NN} & 0.503 & 0.490 \\
\hline
\textbf{GB} & \textbf{0.618} & 0.599 \\
\hline
\end{tabular}
\caption{Experiment 1 - Accuracy (higher is better)}
\label{exp1acc}
\end{center}
\end{table}

\vspace{-0.5cm}

\begin{table}[htbp]
\begin{center}
\begin{tabular}{|c|c|c|}
\hline
\textbf{Classifier} & \textbf{Proposed Approach} & \textbf{Baseline \cite{hasan2009automatic}} \\
\hline
\textbf{DT} & 1.773 & 1.774 \\
\hline
\textbf{KNN} & 2.123 & 2.059 \\
\hline
\textbf{LR} & 1.359 & 1.411 \\
\hline
\textbf{NB} & 3.573 & 3.438 \\
\hline
\textbf{RF} & 1.167 & 1.073 \\
\hline
\textbf{SVC} & 1.358 & 1.426 \\
\hline
\textbf{NN} & 1.204 & 1.353 \\
\hline
\textbf{GB} & \textbf{0.919} & 1.029 \\
\hline
\end{tabular}
\caption{Experiment 1 - MSE (lower is better)}
\label{exp1mse}
\end{center}
\end{table}

\subsubsection{Experiment 2}
The second experiment consists in the classification of Wikipedia articles w.r.t. the seven quality classes by considering, in turn, only the features belonging to each of the three groups in which they can be categorized, i.e., \textit{Text Features} (TF), \textit{Review Features} (RF) and \textit{Network Features} (NF). This experiment aims to identify which group of features is the most discriminating one.

In Table \ref{exp2.1acc} and Table \ref{exp2.1mse} the comparison between the accuracy and MSE values obtained with respect to each group of features are respectively reported.

\begin{table}[htbp]
\begin{center}
\begin{tabular}{|c|c|c|c|}
\hline
\textbf{Classifier} & TF & RF & NF \\
\hline
\textbf{DD} & 0.378 & 0.316 & 0.297 \\
\hline
\textbf{KNN} & 0.415 & 0.287 & 0.303 \\
\hline
\textbf{LR} & 0.469 & 0.395 & 0.332 \\
\hline
\textbf{NB} & 0.3 & 0.255 & 0.198 \\
\hline
\textbf{RF} & 0.502 & 0.391 & 0.373 \\
\hline
\textbf{SVC} & 0.46 & 0.378 & 0.331 \\
\hline
\textbf{NN} & 0.481 & 0.392 & 0.373 \\
\hline
\textbf{GB} & 0.514 & 0.393 & 0.354 \\
\hline
\end{tabular}
\caption{Experiment 2.1 - Accuracy (higher is better)}
\label{exp2.1acc}
\end{center}
\end{table}

\begin{table}[htbp]
\begin{center}
\begin{tabular}{|c|c|c|c|}
\hline
\textbf{Classifier} & TF & RF & NF \\
\hline
\textbf{DD} & 2.085 & 3.013 & 3.607 \\
\hline
\textbf{KNN} & 2.101 & 4.396 & 3.956 \\
\hline
\textbf{LR} & 1.481 & 2.797 & 3.582 \\
\hline
\textbf{NB} & 3.467 & 6.413 & 9.055 \\
\hline
\textbf{RF} & 1.352 & 2.204 & 2.790 \\
\hline
\textbf{SVC} & 1.522 & 3.162 & 3.846 \\
\hline
\textbf{NN} & 1.284 & 2.665 & 2.854 \\
\hline
\textbf{GB} & 1.171 & 2.182 & 2.968 \\
\hline
\end{tabular}
\caption{Experiment 2.1 - MSE (lower is better)}
\label{exp2.1mse}
\end{center}
\end{table}

\vspace{-0.5cm}
As it emerges from both tables, \textit{Network Features} apper to be the less effective, while \textit{Text Features} provide the best level of Accuracy and MSE by employing Gradient Boosting. In particular, TF+GB provides an Accuracy value of $51.4\%$, and an MSE value of $1.171$.

%Figure \ref{features_comparison_rf} shows the comparison between the mean precision, recall and F1-score values obtained by \textit{Random Forest}. As you can see, \textit {Text Features} and \textit{Review Features} are also equivalent according to these metrics, while \textit{Network Features} allow to obtain results much lower than the other two types of features

% \begin{figure}
%     \centering
%     %\includegraphics{}
%     \caption{Caption}
%     \label{fig:my_label}
% \end{figure}

In \cite{bassani18}, other experiments have been provided with respect to other baselines and with respect to the the negative impact that possible noise in the generation of the labeled dataset can have on the classification task.

\section{Conclusions}

In this paper, the problem of assessing the quality of Wikipedia articles has been considered. The collaborative nature at the basis of Wikipedia, where everyone can freely contribute to the creation or modification of articles, represents both an advantage but also a possible drawback of the platform. In fact, if from the one side this could guarantee heterogeneity of sources, skills, and control by a large number of contributors, from the other side it could result in the proliferation of unverified and low-quality contents. %In fact, although within Wikipedia there are policies for controlling the editorial process, based on the activity of teams of volunteers, the growth rate of contents does not always make possible a constant and timely control by human assessors.

For this reason, in the last years, automatic approaches for the classification of Wikipedia contents with respect to given quality classes have been proposed. Most solutions are based on supervised learning techniques, employing multiple kinds of features connected to different aspects of the articles and their authors. With respect to state-of-the-art approaches, in this paper a solution considering a higher number of features has been proposed. The choice of the features is based on an in-depth analysis that encompass the syntax, the style and the editorial history of Wikipedia articles, as well as on a deep investigation of the way in which Wikipedia articles are labeled by WikiProject teams with respect to quality. The promising results obtained confirm the effectiveness of the proposed feature analysis and the interest in continuing the study of the problem.

%In the future, possible research directions could concern a more in-depth study of the importance that single features (or subsets of features) have on the final classification process; furthermore, technological solutions to augment the efficiency in computing Network Features could be proposed. %Finally, it would be interesting to propose methodologies to maintain the alignment between the quality labels provided by the experts and the constant changes that the articles undergo during the editorial process.

%%%%%%%%%%%%%%%%%%%%%%%%%%%%%%%%%%%%%%%%%%%%%%%%%%%%%%%%%%%%%%%%%%%%%%%%%%%%%%%%

\end{document}